\documentclass[letterpaper, 10 pt, conference]{ieeeconf}  % Comment this line out
\IEEEoverridecommandlockouts                              % This command is only
\usepackage{hyperref}
\usepackage{algorithm}
\usepackage[noend]{algpseudocode}
\usepackage{graphicx}
\usepackage{todonotes}
\usepackage{pgfplots}
\usepackage{sfmath}
\usepackage{amsmath}
\usepackage{url}
\pgfplotsset{compat=newest}
\usetikzlibrary{pgfplots.statistics}

\title{\LARGE \bf
Predictive Fault Tolerance for Autonomous Robot Swarms
}
\author{James O'Keeffe, Alan G. Millard% <-this % stops a space
\thanks{J. O'Keeffe and A. G. Millard are with the Department of Computer Science,
        University of York, York, UK
        {\tt\small james.okeeffe@york.ac.uk$^{*}$}}%
}

\begin{document}

\maketitle
\thispagestyle{empty}
\pagestyle{empty}

%%%%%%%%%%%%%%%%%%%%%%%%%%%%%%%%%%%%%%%%%%%%%%%%%%%%%%%%%%%%%%%%%%%%%%%%%%%%%%%%
\begin{abstract}
Active fault tolerance is essential for robot swarms to retain long-term autonomy. Previous work on swarm fault tolerance focuses on reacting to electro-mechanical faults that are spontaneously injected into robot sensors and actuators. Resolving faults once they have manifested as failures is an inefficient approach, and there are some safety-critical scenarios in which any kind of robot failure is unacceptable. We propose a predictive approach to fault tolerance, based on the principle of preemptive maintenance, in which potential faults are autonomously detected and resolved before they manifest as failures. Our approach is shown to improve swarm performance and prevent robot failure in the cases tested.
\end{abstract}

%%%%%%%%%%%%%%%%%%%%%%%%%%%%%%%%%%%%%%%%%%%%%%%%%%%%%%%%%%%%%%%%%%%%%%%%%%%%%%%%
\section{INTRODUCTION}

Autonomous robot swarms are suited to tasks that are dangerous and/or cover large areas because of their multiplicity and redundancy of hardware \cite{csahin2004swarm}. These characteristics were initially thought to provide swarm robotic systems with an innate robustness -- i.e. the ability to tolerate faults and failures. Whilst this is true in some cases, there are others in which faults in individual robots can severely disrupt overall swarm performance \cite{winfield2006safety}. This is especially true for partial failures that do not prevent the afflicted robot from communicating and attempting to interact with the swarm. An active approach to fault tolerance in robot swarms is therefore necessary for achieving long-term autonomy \cite{bjerknes2013fault}. 

Active fault tolerance in robot swarms is generally understood to consist of fault detection, diagnosis and recovery (FDDR) \cite{millard2016exogenous} \cite{o2018fault}. Previous work has examined individual elements of FDDR by spontaneously injecting sensor and actuator faults into individual robots \cite{o2018fault} \cite{tarapore2019fault}
\cite{millard2016exogenous}. However, robots do not tend to spontaneously fail in the field. Rather, failures often result from gradual wear and degradation on sensor and actuator hardware \cite{CarlsonMTBF}. Previous work towards FDDR in swarms has focused on handling faults \textit{after} they have occurred. Reactive FDDR relies upon the assumption that a fault can be resolved during operation -- either autonomously, or by a human in the loop -- and that the swarm can continue operating. There are many instances in which this is not true, particularly in environments that are inaccessible, dangerous, or in enclosed spaces that become quickly congested. Carlson et. al. \cite{CarlsonMTBF} highlight that robots have a mean-time-between-failures -- around 8 hours in the robots they studied. Detecting and handling faults \textit{before} they manifest as failures is advantageous for minimising disruption.

In this paper we demonstrate a novel approach to FDDR in robot swarms whereby faults are \emph{predicted}, allowing at-risk robots to reach a safe area to receive maintenance before the fault manifests as a failure. The contribution of this research is the first predictive FDDR (PFDDR) system for robot swarms and the first such work to consider gradual hardware degradation, as opposed to spontaneous electro-mechanical faults.

\section{RELATED WORK}

\subsection*{Fault Detection}
Fault detection in robot swarms has been approached in a number of ways. Christensen et al. \cite{christensen2009fireflies} use a firefly-inspired approach to communicate the presence of faulty robots to the swarm ; Millard compares a robot's observed behaviour against a simulated model \cite{millard2016exogenous}; Khadidos et al. compare multiple observations of a robot's state with its neighbours \cite{khadidos2015exogenous}; Tarapore et al. use an immune-inspired detection model \cite{tarapore2019fault}; Lee et al. focus instead on determining the most effective metrics by which faults can be detected \cite{lee2022data}. All of these works consider reactive fault detection of spontaneous faults in individual robots.
\subsection*{Fault Diagnosis}
To our knowledge, the only research to address fault diagnosis in robot swarms has been our own immune-inspired work in simulation \cite{o2018fault} and in hardware \cite{o2023hardware}. This approach diagnoses newly detected faults by their statistical similarity to previously resolved ones, but is a reactive approach that relies on the assumption that robots can repair each other in the field.
\subsection*{Fault Recovery}
Khadidos et al. \cite{khadidos2015exogenous} implement recovery by simply powering down faulty robots, becoming inanimate objects for the remainder of a task. Whilst this is acceptable in some scenarios, the accumulation of obstructive objects in tightly enclosed spaces will be problematic if critical paths are blocked. Oladarin \cite{oladiran2019fault} considers a range of different recovery actions for different types of fault, but assumes that robots can repair each other in the field. Christensen et. al. \cite{christensen2009fireflies} use a symbolic recovery mechanism in which the faulty robot is gripped by another robot in the swarm, after which the fault is assumed to be resolved. Bossens and Tarapore \cite{bossens2021rapidly} adopt an alternative approach to fault recovery by evolving a repertoire of robot controllers for a swarm foraging task. If a fault or perturbation is problematic for the current controller, the swarm can then adapt and select a different controller that is less susceptible to the fault, although the fault in the individual is not itself resolved.
\subsection*{Summary}
The existing body of research on swarm fault tolerance is relatively slim. To the best of our knowledge, the literature cited in this section represents the key approaches developed thus far, each of which \emph{reacts} to spontaneous electro-mechanical failures. The approaches that incorporate a recovery or resolution strategy assume that this can be achieved in the field autonomously, which may not always be possible. Thus, there is room for a novel approach that instead \emph{predicts} when robots are at risk of developing faults, such that they are able to preemptively return to a safe area where they can receive maintenance.
\section{METHODS}
This work aims to provide proof-of-concept for new approaches to swarm fault tolerance in software simulation that will translate to near-term hardware solutions. We have chosen GPS-denied underground excavation as a case study application, as it is a topic of interest to industry, the academy, and learned societies around the world \cite{raengictopics2022}. The task is  similar to foraging, which is a common benchmark application in swarm research \cite{bayindir2016review}, but differs insofar that robots manipulate and alter the topography of their environment in real time, navigate tightly constrained spaces as the tunnel is excavated, and must maintain an unbroken communications chain with a known reference point in order to localise. Underground excavation is also a scenario that lends itself to gradual degradation of sensor and actuator hardware due to the build up of dust and debris, thus providing a realistic context for testing our PFDDR approach.

We consider an excavating swarm of robots, in which each robot must remove material from a soil face and return it to the tunnel entrance, forming a long straight tunnel over time (see \autoref{fig:Swarm pic}). Robots localise with respect to a known reference point at the tunnel entrance. We assess tunneling performance simply by the amount of material excavated and the power expended in doing so. Although the ability of the swarm to maintain tunnel dimensions and axis is relevant, it is beyond the scope of this work. For ease of reading, all symbols used in the paper can be found in \autoref{table:symbols}.

\begin{table*}
\begin{center}
\begin{tabular}{ |p{2cm} | p{7.5cm} |  p{7cm}  |}
\hline
Symbol  &  Meaning  & Value or reference  \\ \hline
$M$ & Denotes median value of array in \autoref{fig:PFDDR_State} &  \\
$\mathcal{N}(\mu,\sigma)$ & Gaussian noise about mean $\mu$ with standard deviation $\sigma$. & $\mu = 0$, $\sigma$ is 10\% of the core value it is added to, not inclusive of additional multipliers \\
$M_R$ & Mass of TurtleBot3. & 1 (unitless) \\
$M_L$ & Mass of TurtleBot3 payload. & $0 \leq M_L \leq M_R$\\
$P_0$ & TurtleBot3 battery capacity & 1 (unitless)\\
$\Delta P$ & Rate of power consumption by TurtleBot3 & \\
$v_{max}$ & Maximum linear velocity of TurtleBot3 wheel. & $0.22m s^{-1}$\\
$d_{max}$ & Maximum ultrasonic transmission range. & $2.5m$\\
$\Delta E_{max}$ & Maximum rate of excavation. & $0.2M_R s^{-1}$\\
$dc_{l,r}$ & Degradation severity coefficient on left and right wheels, respectively. & \\
$dc_s$ & Degradation severity coefficient on sensing. & \\
$dc_E$ & Degradation severity coefficient on excavation. & \\

$D_{l,r}$ & Degradation function on left and right wheels, respectively.  Plotted in \autoref{fig:Swarm pic}C. & $(1+e^{-10(dc_{l,r} + \mathcal{N}(0,0.1dc_{l,r}) + \left( \frac{M_L + M_R}{M_R} \right) - 1.5)})^{-1}$\\
$D_s$ & Degradation function on sensing.  Plotted in \autoref{fig:Swarm pic}C. & $2 - 2e^{-2(dc_s + \mathcal{N}(0,0.1dc_s))}$\\
$D_E$ & Degradation function on rate of excavation.  Plotted in \autoref{fig:Swarm pic}C. & $(1+e^{-10(dc_E + \mathcal{N} (0,0.1dc_E)) - 1.5})^{-1}$\\

$v_{l,r}$ & Linear velocity of left and right TurtleBot3 wheels, respectively. & $v_{max}D_{l,r}$ \\
$d$ & Ultrasonic transmission range. & $d_{max} - D_s$\\
$\Delta E$ & Rate of excavation. & $\Delta E_{max} D_E$\\

$p_{l,r}$ & power consumption multiplier on left and right wheels, respectively. Plotted in \autoref{fig:Swarm pic}C. & $2-e^{-5dc_{l,r}}$ \\
$p_E$ & power consumption multiplier on excavation.  Plotted in \autoref{fig:Swarm pic}C. & $2-e^{-5dc_E}$ \\

$\Delta P_{l,r}$ & Rate of power consumption by left and right wheels, respectively. & $(\frac{M_L + M_R}{M_R}) p_{l,r} 2.2P_0 \times 10^{-3}  + \mathcal{N}(0,2.2P_0 \times 10^{-4}) s^{-1}$\\
$\Delta P_s$ & Rate of power consumption by sensing & $1.67P_0\times 10^{-4} 
 + \mathcal{N}(0,1.67P_0 \times 10^{-5}) s^{-1}$\\

$\Delta P_E$ & Rate of power consumption by excavation. & $p_E(0.02P_0 + \mathcal{N}(\mu=0,\sigma=0.002P_0)) s^{-1}$\\

\hline
\end{tabular}
\end{center}
\caption{Symbols used in this paper. Some of the constants given (e.g. $v_{max}$) were selected for their real-world counterparts. Others were informed selections without a specific reference (e.g. $\Delta P_{l,r}$), or accelerated for ease of experimentation (e.g. $\Delta E$). }
\label{table:symbols}
\vspace{-6mm}
\end{table*}

\subsection{Swarm Robot Model}
We have evaluated the efficacy of our proposed approach in simulation, using swarms of simulated TurtleBot3 robots \cite{TB3}. We do not suggest that the TurtleBot3 is itself a suitable platform for underground excavation, but its availability and Robot Operating System (ROS) integration make it an ideal test platform for producing repeatable proof-of-concept data. 

We provide our simulated robots with the ability to localise other robots within 2.5 metres relative to themselves with a 10\% margin of error, based on the distributed ultrasonic approach verified in hardware by Maxim et. al. \cite{maxim2008trilateration}. Robots can share self-estimated state information with any other robot within 50m -- something that could be achieved with a Decawave 1000 chip \cite{DW1000}, for example.  The robots are also equipped with directional proximity sensors at $\pm$ 0$^{\circ}$, 30$^{\circ}$, 60$^{\circ}$, and 90$^{\circ}$ for collision avoidance. 

Excavation typically requires a drilling or cutting tool, and a container to store and transport excavated material. In-depth modelling of soil excavation is beyond the scope of this work. Rather, each robot is assumed to be able to remove approximately its own volume worth of material from the soil face. For simplicity, we take a loaded robot to weigh twice as much as an unloaded robot.

\subsection{Tunnelling Algorithm}
\begin{figure*}[!tbp]
  \centering  
    \includegraphics[width=0.9\textwidth]{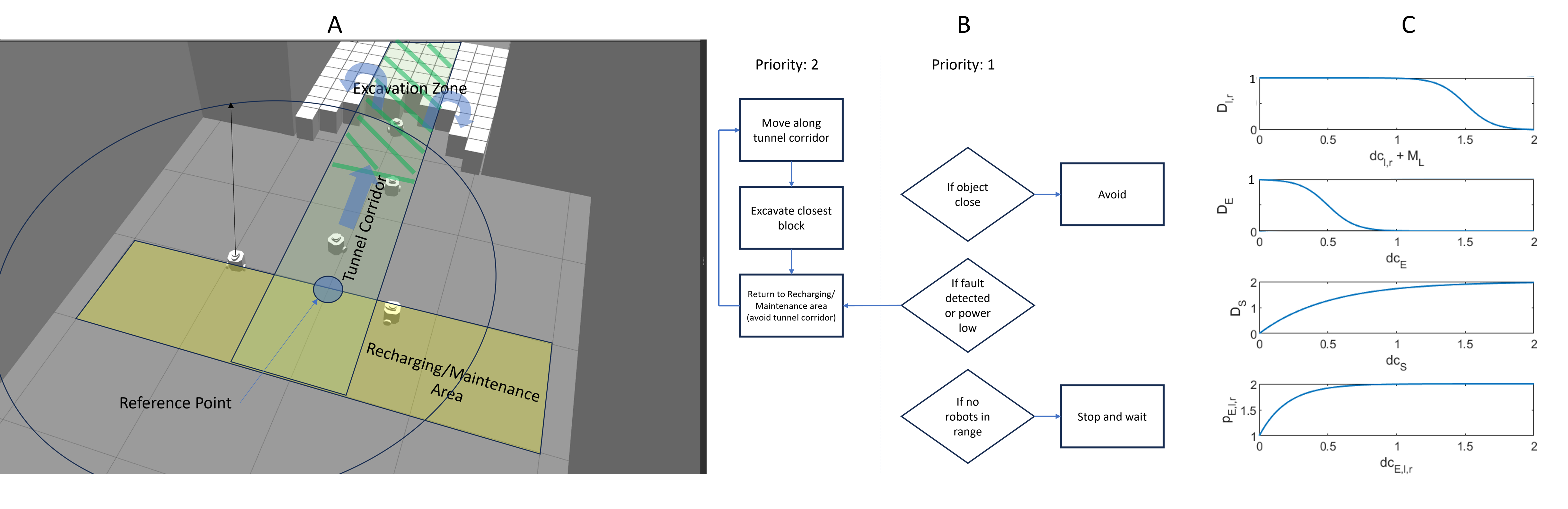}
    \caption{\textbf{A:} A screenshot of the simulated tunnelling swarm with highlighted maintenance and recharging zone, digging corridor, excavation zone, communication range of an individual robot (approximately to scale), and arrows indicating the path taken during the excavation algorithm. \textbf{B:} A high level state machine providing an overview of our proposed excavation algorithm. Avoiding collisions, maintaining an unbroken communication chain, and returning to the recharging/maintenance area when needed will take priority over the basic excavation algorithm. \textbf{C:} Relationships between degradation functions and degradation severity coefficients listed in \autoref{table:symbols}.}
    \label{fig:Swarm pic}  
\end{figure*}

We propose a naive swarm tunnelling algorithm, exploiting local sensors and actuators. The algorithm can be summarised as follows: Each robot begins in a demarcated charging and maintenance zone outside of the excavation area. It will then move into a defined tunnel corridor (0.8m across) and head along the tunnel axis away from the charging/maintenance zone and into the excavation zone (defined as the tunnel corridor further than 1.5m from the charging/maintenance zone) until it encounters a soil face, at which point it will enter an excavation state and remove a single block of soil.
Soil is modelled as discrete blocks covering $0.2$m x $0.2$m of ground (we do not consider tunnel height in this work). The robot will then carry the soil back to the recharging and maintenance area, where it is assumed to be deposited (see \autoref{fig:Swarm pic}).

When a robot requires maintenance, or its power drops below 30\%, it will interrupt its activities to return to the charging and maintenance area. Each robot performs collision avoidance for any object within 0.5m and $\pm$ 90$^{\circ}$. The swarm maintains a chain-link of communication with the tunnel reference point by ensuring that each robot stays within 2m of its nearest neighbour with an unbroken communication link to the tunnel reference. To avoid clustering in enclosed spaces, a robot headed towards the soil face will maintain a distance greater than 1m from any robot it detects as being further along the tunnel than itself.

\subsection{Power Consumption and Degradation Models}
In this work, there are three robot processes that consume power and are subject to degradation at different rates: 1) sensing and communication, 2) locomotion, and 3) excavation. We use a naive model of power consumption whereby each process consumes some percentage of total battery capacity, $P_{0}$, per unit time. We simulate the susceptibility of these processes to degradation caused by dust and debris thrown into the air from the excavation process, and from traversing loose ground. Degradation affects both the physical process and the power consumed by it. For example, debris accumulating on motor hardware will create resistance as the motor turns, slowing the motor, generating heat, increasing  power consumption, and reducing efficiency. We assume that, as with most electrical devices, the average power drawn by each actuator will be considerably lower than the maximum possible power draw -- typically around 50\%. As hardware degrades, the physical effects will initially not be noticeable, as there will be some overhead to draw more power to meet the desired output. However, as degradation continues, the power drawn by an actuator will increase to its limit, at which point degradation will manifest physically -- e.g. as a reduction in velocity or rate of excavation. 
% The degradation functions are shown in \autoref{fig:Swarm pic}B-E.
The power consumption and degradation models for each process is as follows:
\subsubsection*{Sensing and Communication} Each robot is constantly emitting and receiving simulated ultrasonic signals and sharing state information. The power consumed by sensing hardware, $\Delta P_s$, is unaffected by the accumulation of debris on sensor hardware. However,
the degradation function, $D_s$ affects the ultrasonic transmission range used for robot localisation. We implement this such that a robot's sensing range can drop to a minimum of 0.5m.
\subsubsection*{Differential Drive} The power consumed by robot locomotion will depend on the robot's payload, $M_L$, and will be affected by the amount of dust and debris accumulated on wheels and motors, modelled as the multiplier $p_{l,r}$, a function of degradation coefficient $dc_{l,r}$, on $\Delta P_{l,r}$ for left and right wheels, respectively. We assume that a robot is able to transport its maximum payload without non-linear effects. 
Degradation also affects the velocity of each wheel, $V_{l,r}$. An unloaded robot will not suffer any reduction in velocity until degradation severity coefficient $dc_{l,r} > 1$ because the increased power draw from the degradation function, $D_{l,r}$, in this range would be no greater than the increased power draw from carrying its extra payload.
\subsubsection*{Excavation} This work uses a simplified representation of the excavation process, in which discrete blocks are removed once a robot has been in an excavating state for an appropriate amount of time. The power consumed by the excavation process, $\Delta P_E$, is affected by multiplier $p_E$, a function of degradation severity coefficient $dc_E$. 
The rate of excavation, $\Delta E$, is affected by degradation function, $D_E$.
Under ideal conditions ($dc_E = 0$), removing the maximum volume of soil that a robot can carry (i.e. $M_L = M_R$) will expend 10\% of a robot's total battery capacity.
\subsubsection*{Recharging} For ease of experimentation, and because the rate of recharging does not directly affect the metrics by which we measure the quality of our fault tolerance system, we accelerate the rate of robot recharging to 10\% per second.

\section{IMPLEMENTATION, RESULTS \& DISCUSSION}
\subsection{Normal and Faulty Behaviour Baselines}

All experiments were conducted using ROS 2 (Foxy) and Gazebo Classic. 10 replicates were performed for each experiment. We first tested the performance of the tunnelling algorithm under ideal conditions (no degradation/faults), using a swarm of five robots. This is a relatively small size for a swarm, however the number of robots is proportional to tunnel dimensions. All algorithms presented are decentralised and scalable in principle with longer/wider tunnels (to be confirmed in future work).
We tested how each type of fault would affect swarm performance, depending on how many robots were afflicted. In these experiments, each fault type is considered in isolation, with the number of faulty robots varying between 0-5. Every faulty robot has a fixed 15\% probability that its associated degradation severity coefficient, $dc_{S,E,l,r}$, will increase by an increment of 0.01 per second of simulated time. We assess swarm performance by the total number of blocks excavated by the swarm, and the power consumed in doing so. 

\begin{figure*}[!tbp]
  \centering  
    \includegraphics[width=\textwidth]{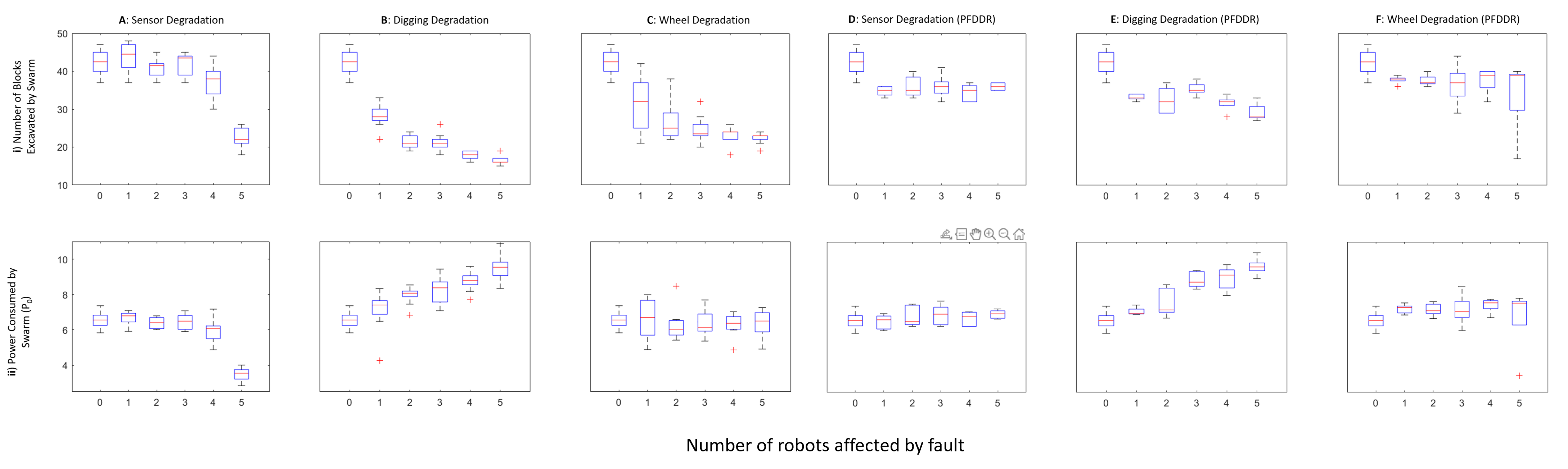}
     \vspace{-5mm}
    \caption{Row \textbf{i} shows the total number of blocks excavated by the 5-robot swarm in 15 minutes of simulated time, when up to 5 robots suffer from \textbf{A}: sensor degradation, \textbf{B}: excavation degradation, or \textbf{C}: motor degradation without our PFDDR system; or \textbf{D}: sensor degradation, \textbf{E}: excavation degradation, or \textbf{F}: motor degradation with our PFDDR system. Row \textbf{ii} shows the total power consumed by the 5-robot swarm in 15 minutes of simulated time as a percentage of a single robot's battery capacity, where columns \textbf{A-F} indicate the same fault categories, with and without our PFDDR system.}
    \label{fig:FaultAnalysis}  
    \vspace{-2mm}
\end{figure*}

\autoref{fig:FaultAnalysis}A shows that sensor degradation in individual robots has the least impact on performance in terms of both power consumption and block removal. This is because of the built-in redundancy of the distributed localisation technique, and because the swarm operates over a relatively small area during the 15 minutes of simulated time per experiment, with excavated tunnel depth only ever reaching a maximum 1.6m. The swarm therefore never needs more than one unaffected robot acting as a chain link to the reference point at any time. In all cases where 5 robots were affected by sensor degradation, the swarm completely lost the ability to localise itself before the end of the experiment. This suggests that a comparable degradation in overall performance could be expected with fewer affected robots were the swarm to be distributed over a larger physical area. 

\autoref{fig:FaultAnalysis}B shows that degradation in excavation hardware substantially reduces the number of blocks excavated and increases power consumption, with robots eventually becoming completely incapable of excavating, and hindering the rest of the swarm from reaching the soil face until returning to the recharging area. Because of the 30\% cut-off at which a robot will interrupt its current activity to seek charge, robots with degraded digging hardware in these experiments always manage to reach the recharging zone before completely depleting their charge.

\autoref{fig:FaultAnalysis}C shows that degradation of motor hardware also significantly reduces the amount of blocks that the swarm can excavate in an experiment. Although motor degradation does not appear to have much impact on the total power consumed by the swarm, this is partly because an average robot operating in ideal conditions only needs to replenish its power once in 15 minutes of simulated time. This obscures the fact that robots with motor degradation eventually become incapable of moving and completely deplete their power, unable to reach the recharging and maintenance zone. Every robot with unaddressed motor degradation eventually reaches this state.

\subsection{Predictive Fault Tolerance}

The physical effects of degradation are negligible for $dc_E < 0.3$ and $dc_{l,r} < 0.3$, with the main impact seen in rate of power consumption which approximately doubles (see \autoref{table:symbols}). As a worst-case scenario, the system should therefore aim to detect degradation on excavating and motor hardware before $dc_E, dc_{l,r} > 0.3$. Given the relatively small impact that sensing degradation has on the system, understanding when it should be detected and resources expended on maintenance is less obvious. The root cause of failure occurs when sensing range drops below 2m, a hard coded value for maintaining proximity. As a general rule, then, a robots sensing range should never be allowed to drop below the greatest value hard coded into its controller. We implement our PFDDR system as follows:
\subsubsection*{Detection and Diagnosis}
\begin{figure}[!tbp]
  %\centering  
    \includegraphics[width=0.48\textwidth]{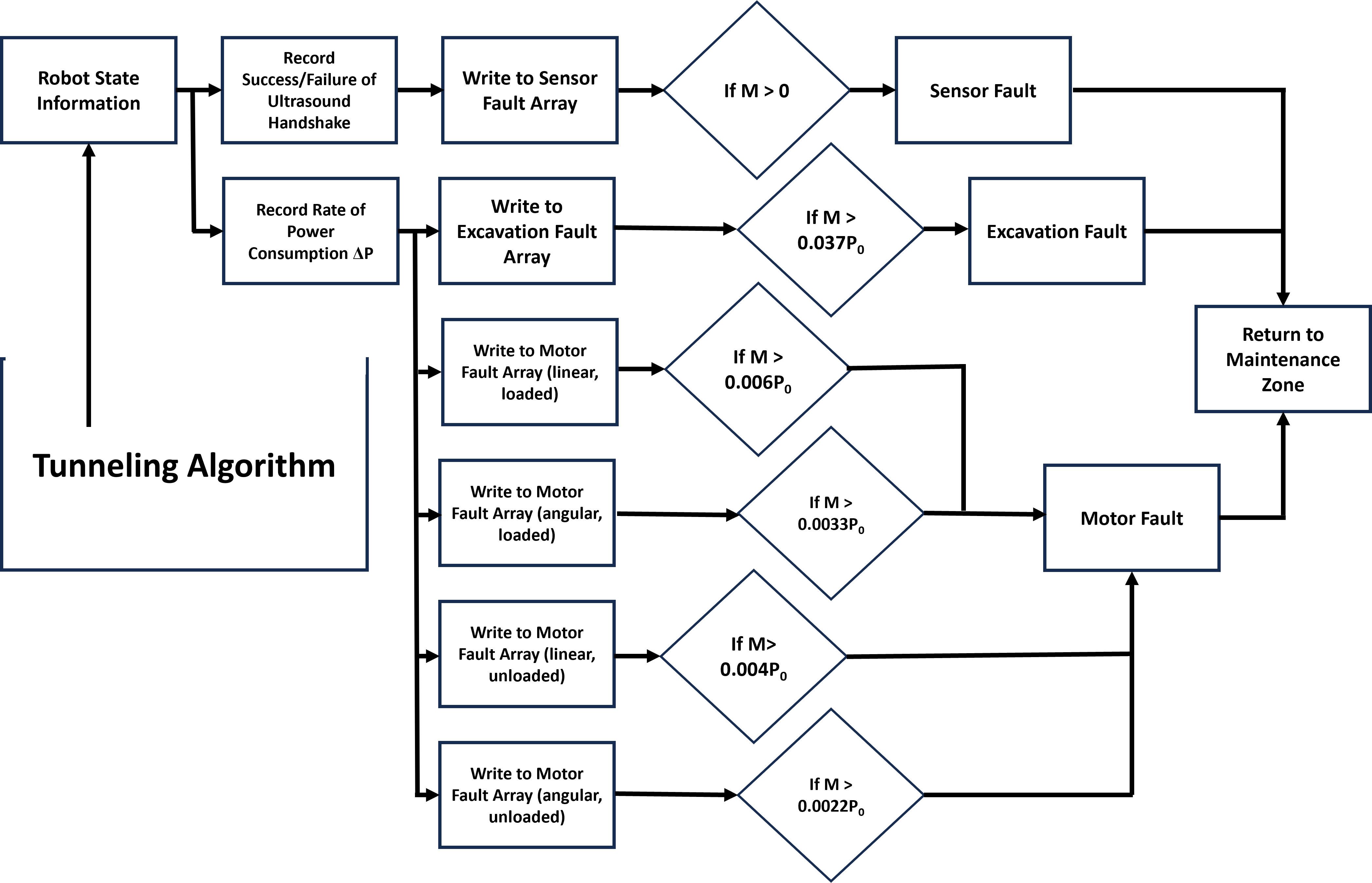}
    \caption{A high level state machine outlining our proposed PFDDR system. Each constant is selected with respect to the equations given in \autoref{table:symbols} and the rates of power consumption observed during experiments for $dc_E < 0.3$ and $dc_{l,r} < 0.3$. }
    \label{fig:PFDDR_State}  
\end{figure}

In this scenario, our system must detect and differentiate between 3 categories of fault: sensor hardware, motor hardware, and excavation hardware. The orthogonality of these categories allows us to hard-code diagnosis according to the data sets in which faults are detected. Each type of fault category is assigned an array into which robot states are recorded. Robot states are only recorded when a robot is performing the appropriate corresponding task -- i.e. a robot that is excavating will write its power consumption state to a separate array than if it had been travelling. For motor faults, power consumption varies significantly depending on whether the robot is using one or both wheels, and whether it is carrying a payload. There are therefore four separate arrays for monitoring power consumed by locomotion. Each robot controller updates at a rate of 100Hz. Every state recorded into the array is averaged over 10 -- i.e. states are recorded at a rate of 10Hz. Each array can contain up to 50 states, meaning that a full array represents 5 seconds of simulated time, although not necessarily consecutive. A fault is detected when the median value of its associated array is greater than a threshold value. For motor and excavation faults, robots record their power consumption states. For sensor faults, a handshake protocol is used whereby each robot will listen for confirmation that detected neighbouring robots have also received the simulated ultrasonic emission from the robot. If any neighbour within 2m does not confirm, the robot writes a 1 to the array, otherwise it writes a 0. If the median value of the sensing array is greater than 0, a sensor fault is detected. This process is shown in \autoref{fig:PFDDR_State}.

\subsubsection*{Recovery}
A robot returns to the recharging and maintenance area \autoref{fig:Swarm pic}. Here, a robot will receive maintenance on the sensors/actuators that have been flagged in the detection/diagnosis process. It is assumed that maintenance is performed by a human, although it could be performed autonomously in the near-mid future. Once a robot has received maintenance on a given sensor or actuator, its corresponding degradation severity coefficient is set to zero. Performing maintenance on a sensor/actuator takes 5 seconds of simulated time (accelerated for ease of experimentation).

\subsection{Faults in Isolation}
To test our PFDDR system, we reran our experiments with each fault type in isolation whilst the system was active.

\autoref{fig:FaultAnalysis}D-F show that the power consumed by the swarm when our PFDDR system is implemented is largely the same as without. For sensor faults, there is a slight decrease in the number of blocks excavated when our PFDDR system is active. This can be explained by the fact that robots will spend more experiment time returning to the maintenance area when faults are detected. In the case of motor and excavation faults, the total number of blocks excavated by the swarm is significantly improved by our PFDDR system.

\subsection{Faults in Combination}

We next experiment with all fault types in combination. As it is unlikely in a real world scenario that all robots in a swarm would degrade at the same rate, we initialise each fault type with a random probability (between 1\% and 15\% per second of simulated time) of its corresponding degradation severity coefficient increasing by 0.01. 

\begin{figure}[!tbp]
  %\centering  
    \includegraphics[width=0.48\textwidth]{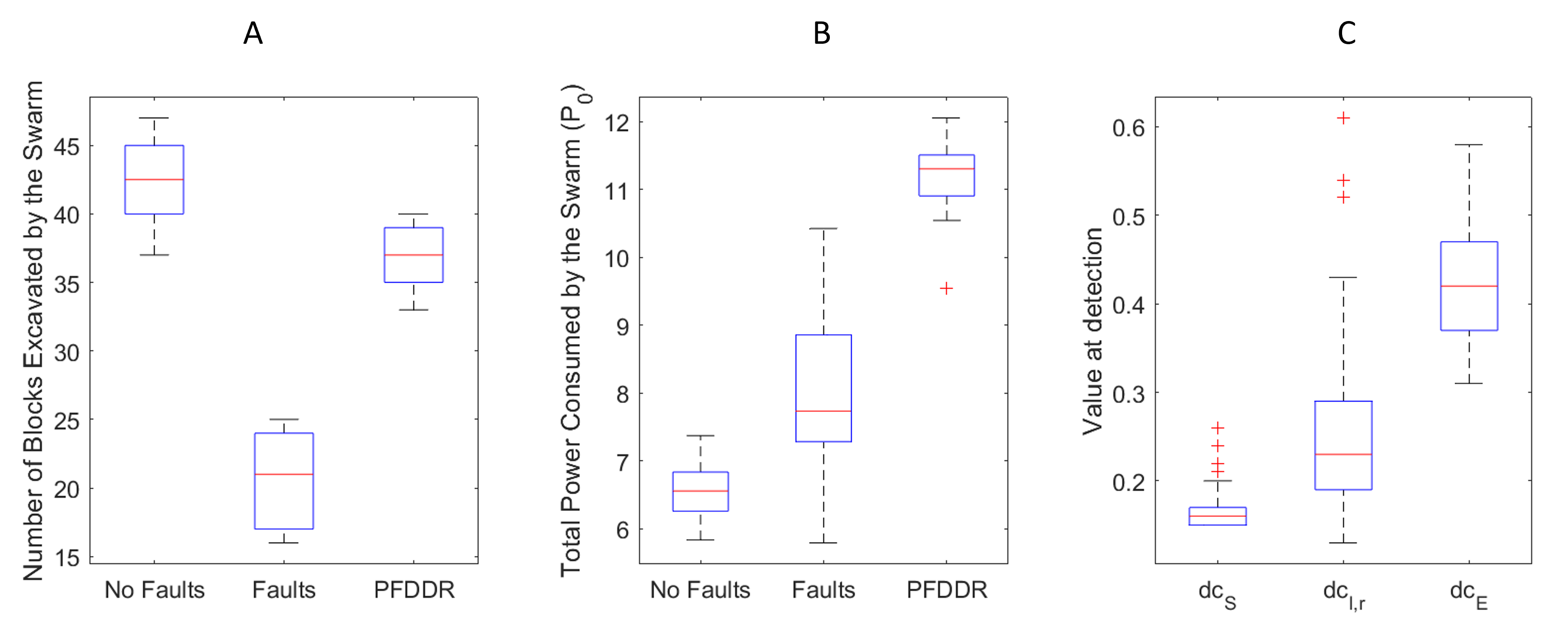}
    \caption{\textbf{A} shows a comparison in the total number of blocks removed by the swarm when the swarm operates in ideal conditions, with faults in combination and without the implementation of our predictive PFDDR system, and with faults in combination with the implementation of our PFDDR system. \textbf{B} shows a comparison in the total power consumed by the swarm as a percentage of the total battery capacity of a single robot when the swarm operates in ideal conditions, with faults in combination and without the implementation of our PFDDR system, and with faults in combination with the implementation of our PFDDR system. \textbf{C} shows the value of degradation severity coefficients at the time a fault was detected.}
    \label{fig:FDDRAnalysis}  
\end{figure}

\autoref{fig:FDDRAnalysis}A and B show the performance of the swarm with and without the PFDDR system, with the swarm operating in ideal conditions included as a control. Unsurprisingly, overall performance is vastly reduced when robots are left to degrade over the course of the experiment. Performance improves considerably when our PFDDR system is active. 

\autoref{fig:FDDRAnalysis}A shows that, when our PFDDR system is implemented, the total number of blocks excavated by the swarm is far closer to the ideal conditions control than when the swarm is allowed to degrade. Across all 10 experimental replicates, the total number of robots that deplete their entire power supply after 15 minutes of simulated time when the system is left to degrade is 9. That number is reduced to 0 when our PFDDR system is implemented. Whilst we would not expect our prototype PFDDR system to completely eliminate robot failure in all cases as is, this result serves as a strong validation of the principle of PFDDR in preventing robot failures. A reactive FDDR system, by comaprison, would not detect a fault until a later stage of degradation. In the case of motor failure, this would potentially require a faulty robot to be retrieved from the tunnel, causing severe disruption. A direct comparison of PFDDR vs. reactive FDDR will be included in future work.
\autoref{fig:FDDRAnalysis}B shows that the total power consumed by the swarm when our PFDDR system is implemented is significantly higher than when the swarm is left to degrade, and almost twice as high as our ideal conditions control. To some extent, this is unsurprising as a a robot that is left to degrade will eventually fail to reach the recharge/maintenance zone and cease to consume power once its remaining supply is depleted.

Nonetheless, \autoref{fig:FDDRAnalysis}B demonstrates a hitherto unreported artifact of swarm fault tolerance that, if the rate of power consumption is the first indication of hardware degradation, a degraded system will inevitably consume more power than an ordinary system model would account for -- even if active PFDDR prevents it from ever manifesting as a failure. Assessment of power consumption should therefore be included in future analysis of the effectiveness of swarm PFDDR systems. Given that there are many scenarios in which charging resources might be limited, this result is open to criticism as a negative feature of our system and warrants further investigation. However, we would argue that, for autonomous systems in general, consuming additional power sources is preferable to failure at a critical moment. Furthermore, the severity of the result is very likely due to the simple means of fault detection, for which there is considerable scope to improve. 

\autoref{fig:FDDRAnalysis}C displays the values of each type of degradation severity coefficient at the moment our PFDDR system detects a fault. Sensor faults are typically detected very quickly ($dc_S = 0.16$ on average). Motor faults tend to be detected whilst $dc_{l,r} < 0.3$, but occasionally at much higher values -- usually because the degradation severity coefficient has increased whilst robot has been stationary and therefore not recording power consumption states. This effect is far more obvious in the detection of excavation degradation, which typically occurs at $dc_E > 0.4$ -- considerably higher than desired. This is because robots spend a minority of time in the excavation state, whilst $dc_E$ gradually increments unnoticed. This highlights a need for active precautions against passive degradation -- for example, periodic diagnostic checks similar to those implemented in our previous work \cite{o2017diagnosis} could be used to routinely check for degradation on actuators that are used less frequently.

\section{CONCLUSIONS \& FUTURE WORK}

In this paper we have argued that swarm fault tolerance should be predictive, as there are real world scenarios in which any robot failure is unacceptable.
We consider a swarm excavation scenario and propose a PFDDR system. We demonstrate that PFDDR can be used avoid robot faults manifesting as failures by detecting and diagnosing early signs of degradation and performing preemptive maintenance. Although preemptive maintenance is not itself a new concept, we are the first to apply it to swarm robotics. Our system was able to maintain a comparable rate of excavation and completely prevent the loss of robots in the cases tested, albeit at a significantly increased power cost.
Whilst the system described here is a simple proof-of-concept, this work represents a fundamental shift in approach to swarm fault tolerance, as well as multi-robot systems in general.

There are many further avenues to explore within this approach, and it is our hope that other swarm researchers will adopt a predictive approach in their own work on fault tolerance and build on our progress. Our future work will investigate more sophisticated approaches to predictive detection and diagnosis (e.g. machine learning), incorporate path-planning algorithms into robot behaviour and recovery strategies, and use hardware experiments to produce high fidelity models of power consumption and hardware degradation. We will also compare our PFDDR approach against state-of-the-art FDDR approaches.

\subsection*{Limitations}
\subsubsection*{Naive Models}
We use naive models of power consumption and hardware degradation, as well as a naive controller algorithm as the testbed for this work. Whilst our system implementation provides effective proof-of-concept for our predictive FDDR system, we acknowledge that our models will not transfer directly into a real world system without modification.

\subsubsection*{Simulated Data}
We are precluded from performing genuine hardware experiments by the fact that robots capable of autonomous excavation are not commercially available. We will improve our models in future work, for example by measuring the power consumption of the types of robot platforms that are powerful enough to excavate (e.g. Clearpath models) and of drill augers. Our previous work on swarm fault diagnosis shows that trends observed in simulation tend to be replicated in hardware \cite{o2023hardware}.

\section*{ACKNOWLEDGEMENTS}
This work was funded by the Royal Academy of Engineering UK IC Fellowship Award, ICRF2223-6-121 (O'Keeffe).


\bibliographystyle{plainurl}
\bibliography{main.bib}

\end{document}